\documentclass[letterpaper, 10 pt, conference]{ieeeconf}

\IEEEoverridecommandlockouts

\usepackage{graphicx}
\usepackage[caption=false, font=footnotesize]{subfig}
\usepackage{algorithm}
\usepackage{algpseudocode}
\usepackage{comment}
\usepackage{color}
\usepackage{booktabs}
\usepackage{multirow}
\usepackage{amsmath}
\usepackage{amssymb}
\usepackage{bm}
\usepackage{booktabs}
\usepackage{tikz}
\usetikzlibrary{positioning,calc,fit,arrows.meta}

\title{\LARGE \bf
DGF-Asym: Adaptive Peer Weighting for Collective Estimation \linebreak under Noisy and Faulty Sensing}

\author{Mohammad Nomaan Husain$^{1}$ and Tanja Katharina Kaiser$^{1}$
\thanks{$^{1}$Department of Computer Science \& Artificial Intelligence, University of Technology Nuremberg, Germany {\tt\small nomaan.husain@utn.de}}%
}

\begin{document}

\maketitle
\thispagestyle{empty}
\pagestyle{empty}

\begin{abstract}
Collective estimation under unreliable sensing requires robots to adaptively balance personal observations and social information from peers. 
We propose DGF-Asym, a decentralized collective estimation method extending the DeGroot-Friedkin model with asymmetric peer weighting based on retrospective predictive performance. 
The method supports consensus formation without prior sensor-noise statistics or explicit fault models while keeping inter-robot communication asynchronous. 
We evaluate DGF-Asym in simulated collective navigation under noisy observations and time-varying sensor faults. 
Compared with three baselines, it produces more stable estimates and more reliably sustains the decision criterion, at the cost of longer decision times. 
However, overall performance depends on fault ratio, group size, and influence-threshold selection. 
Our results demonstrate the potential of adaptive peer weighting for stable model-free collective estimation under unreliable sensing.
\end{abstract}

\section{INTRODUCTION}

Multi-robot systems can accomplish tasks that are difficult or impossible for a single robot by collectively making decisions, acting, and exchanging information about their environment. Their effectiveness, however, depends critically on the reliability of the information shared among team members. In open or adversarial settings, robots may receive false or misleading information from external intruders or from compromised or defecting members of the team. Previous work has shown that even small numbers of disruptive robots can substantially impair the performance of an entire multi-robot system~\cite{Strobel2018}. 

This raises a fundamental challenge for robust collective estimation: how should robots combine information received from others with evidence obtained through their own sensors? While communication can increase the amount of available information and thus compensate for limited individual observations, blindly trusting communicated information makes the system vulnerable to misinformation~\cite{Strobel2018}. Conversely, relying on local observations discards potentially valuable knowledge held by other robots. Effective information fusion, therefore, requires mechanisms that balance these two sources of information while accounting for their uncertainty and reliability. In this work, we study information fusion in multi-robot systems in the presence of potentially unreliable information, with the goal of enabling individual robots to form accurate beliefs while remaining robust to faulty observations communicated by other agents. We propose a fully decentralized collective estimation method for multi-robot systems based on the DeGroot-Friedkin model~\cite{Jia_MirTabatabaei_Friedkin_Bullo_2015} that we extend with dynamic peer-specific social-information weighting. DeGroot-Friedkin's self-appraisal mechanism enables robots to continuously update the weight they put on their personal information based on the perceived influence on the collective estimate. Our dynamic peer-specific social-information-weighting mechanism further enables robots to assign different importance levels to information received from individual neighbors, rather than treating all neighbors with the same importance as in the original formulation. We evaluate the method under increasing sensor noise, varying proportions of robots with defective sensors producing random observations, and different group sizes in a realistic robot simulator. In a collective navigation task, we demonstrate that our proposed dynamic peer-specific social weighting extension of the DeGroot-Friedkin model results in improved estimation stability, while being sensitive to fault ratio and group size.  
\begin{figure}
    \centering
    \includegraphics[width=1.0\linewidth]{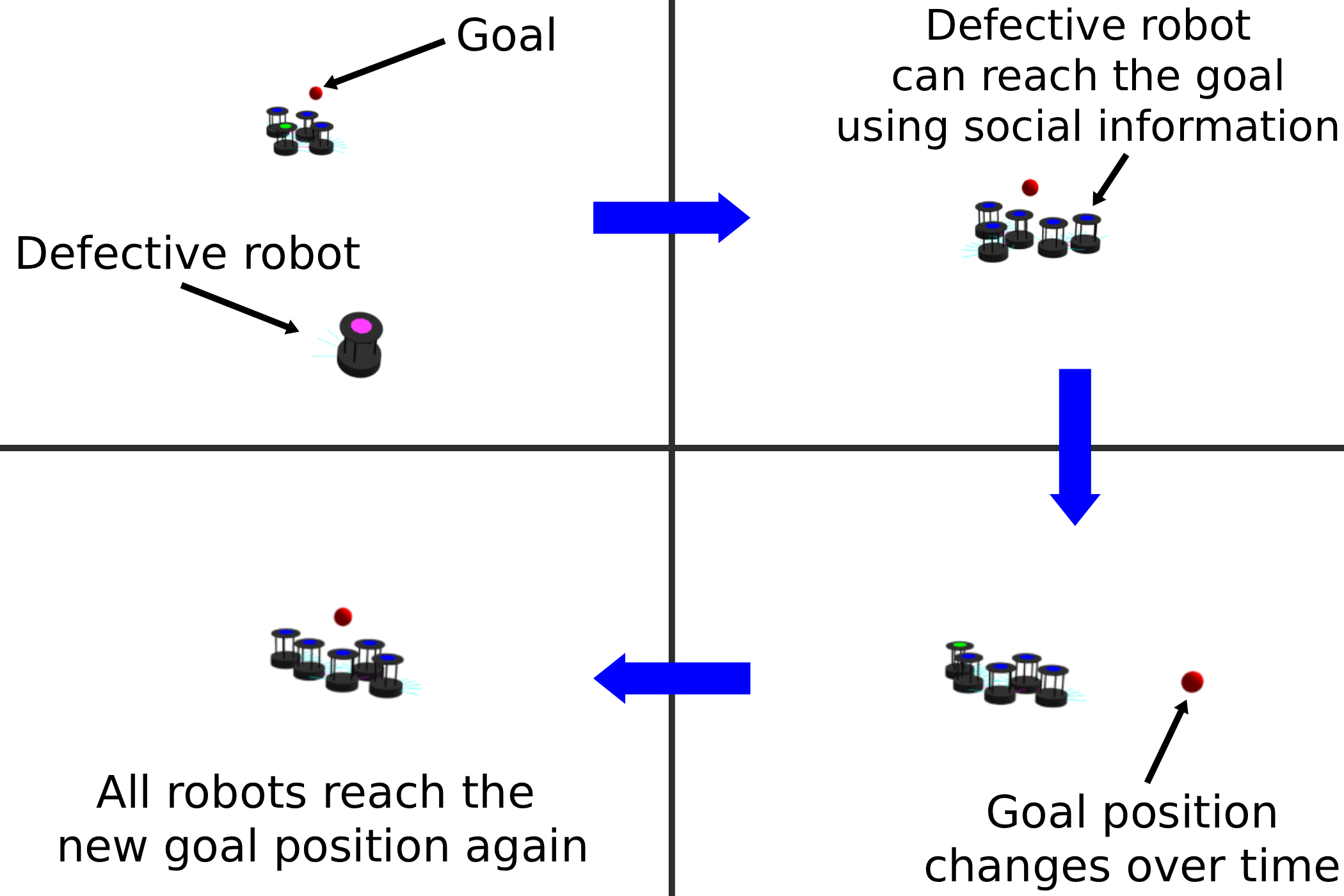}
    \caption{Visualization of our collective estimation task in the collective navigation application scenario.}
    \label{fig:placeholder}
\end{figure}

In summary, we make the following contributions: 

\begin{enumerate}
  \item We propose a dynamic peer-specific social-information weighting method for collective estimation in multi-robot systems, extending the DeGroot-Friedkin model.  
  \item Our method has no probabilistic assumptions about sensor noise and works with asynchronous communication.
  \item A simulation-based evaluation against the standard DeGroot-Friedkin model and two baselines demonstrates improved estimation stability and characterizes trade-offs in accuracy and sustained decision success.
\end{enumerate}

\section{RELATED WORK}
\label{sec:related_work}

\subsection{Bayesian Fusion}

Bayesian fusion combines information from multiple uncertain sources through probabilistic models of states and observations~\cite{Durrant-Whyte2008}.  
Kalman and extended Kalman filtering (EKF) instantiate this principle by weighting model predictions against measurements according to their estimated uncertainties~\cite{gelb1974applied}. 
Standard formulations typically assume that the relevant process and measurement noise statistics are known or specified. 
Adaptive methods reduce this dependence by innovation-based covariance estimation~\cite{mohamed1999adaptive} or variational Bayesian adaptation~\cite{sarkka2013non}. 
Liu et al.~\cite{liu2020fully} extend noise-adaptive estimation to distributed sensor networks by jointly updating local states and measurement-noise precision matrices. 
Although these methods can infer previously unknown noise parameters, they still rely on an explicit probabilistic state and observation model and assumptions about the form of the underlying noise distributions.

\subsection{Collective Perception and Adaptive Information Weighting}

Collective perception enables groups of robots to infer environmental properties from distributed observations~\cite{Valentini2016}. 
While collective decision making selects among discrete alternatives, collective estimation seeks to infer an underlying environmental state or quantity~\cite{ShanMostaghim2021DBBS}. 
Bayesian approaches such as Bayes Bots~\cite{Ebert2020BayesBots} and distributed Bayesian belief sharing~\cite{ShanMostaghim2021DBBS} combine uncertain local and social information, while change-detection methods enable adaptation to dynamic environments~\cite{PfisterHamann2023ChangeDetection}.

A related line of work addresses uncertainty in source reliability. Minimalistic Collective Perception~\cite{ChinKhalufPinciroli2023MCP} incorporates sensor accuracy in collective estimation, while the Adaptive Sensor Degradation Framework~\cite{ChinPinciroli2024ASDF} and BayesCPF~\cite{ChinPinciroli2025BayesCPF} adapt sensor-accuracy estimates online using social information. In contrast, our approach adapts reliance on personal and social information without explicitly estimating sensor accuracy or assuming a sensor-degradation model. 

Reliability has also been represented through trust and adaptive fusion weights. Pierson and Schwager~\cite{PiersonSchwager2016} infer inter-robot trust from relative sensing performance. However, they use it to allocate spatial responsibility rather than contributions to a shared estimate.  Hallyburton and Pajic~\cite{HallyburtonPajic2025} maintain probabilistic trust states for sensor fusion using physical-model consistency and peer information, whereas our approach does not require an explicit trust or consistency model. Gupta et al.'s Autonomous Online Learning (AOL) framework~\cite{Gupta2024AOL} is the closest to our setting, directly adapting local and social fusion weights under sensor failures using online learning signals. Unlike AOL, our weights are derived from interaction dynamics: personal reliance is adapted from perceived social influence, and peer reliance from retrospective predictive performance.

\subsection{Opinion Dynamics}

Opinion dynamics provide a complementary perspective on information fusion by modeling how agents revise their beliefs through repeated social interaction. DeGroot's model~\cite{DeGroot_1974} formalizes weighted opinion averaging, while the DeGroot-Friedkin model~\cite{Jia_MirTabatabaei_Friedkin_Bullo_2015} extends this framework with self-confidence adaptation through reflected appraisal of social influence. Subsequent extensions introduce finite-time, neighbor-based self-appraisal~\cite{Xu2015ModifiedDGF} and issue-dependent interaction topology~\cite{Ye2017DynamicDGF}. These models provide a basis for adapting information weights from interaction dynamics rather than from explicit estimates of sensing uncertainties. 

Building on this perspective, our approach applies influence-based self-appraisal and retrospective peer weighting to collective estimation. Unlike probabilistic fusion and sensor-calibration approaches, the resulting weights represent interaction-derived reliance rather than calibrated measures of sensor accuracy or uncertainty.

\section{METHODS}

\subsection{Problem Description}
A team of $N$ mobile robots operates in a two-dimensional workspace $\mathcal{W}\subset\mathbb{R}^2$. The objective is to reach agreement on a continuous variable $p^\text{gt}(t)$, which may vary over time~$t$, resulting in a collective estimation problem. Each robot~\(i\) obtains a local sensor observation~\(z_i(t)\) of this variable. However, these measurements may be affected by noise, variations in sensing quality, or unidentified faults, causing the reliability of the information available to individual robots to vary over time.

To improve accuracy, each robot monitors its communication channel for broadcasts from neighboring peers containing their current estimates \(\hat{z}_j(t)\). Communication among robots is asynchronous, with each robot \(i\) broadcasting its current estimate at intervals \(\Delta t_i^{\mathrm{tx}}\). Robot \(i\) then forms its own estimate, \(\hat{z}_i(t)\), by combining its local sensor observation with the social information received from its peers.

\subsection{Asymmetric Peer Weighting Model}

We propose DGF-Asym, an asymmetric peer weighting model based on the DeGroot-Friedkin (DGF) model for collective estimation. 
In particular, we introduce a formulation of peer-specific weighting of social information.

\subsubsection{Multi-Robot DeGroot-Friedkin Model}
The DeGroot-Friedkin (DGF) model describes how individuals repeatedly form opinions across a sequence of issues while adapting their self-confidence over time~\cite{Jia_MirTabatabaei_Friedkin_Bullo_2015, Jia_Friedkin_Bullo_2017}. Within each issue, opinions evolve through DeGroot averaging~\cite{DeGroot_1974}, while between issues, self-weights are updated through Friedkin's reflected-appraisal mechanism~\cite{friedkin_2011}, such that greater influence over the previous group outcome leads to greater self-confidence in the subsequent issue.

We adapt the DGF model for collective estimation in a multi-robot system by defining a \textit{sequence of issues} as $k$~update intervals of $T_{\mathrm{up}}=25\,s$ each. The self-confidence of robot $i$ during update interval $k$ is denoted by \(\omega_{ii}(k)\in[\omega_{\min},1-\omega_{\min}]\) with \( 0 < \omega_{\min} \lll 1\). It specifies how strongly robot $i$ weights its own local observation relative to the social information received from its peers. The self-weight~$\omega_{ii}$ is bounded away from the extremes by~$\omega_{\min}$. If all robots reach $\omega_{ii}=0$, new sensory information no longer enters the collective estimate, rendering the system unresponsive to environmental changes. Conversely, at $\omega_{ii}=1$, a robot disregards social information and can no longer be corrected by its peers.

Each robot~\(i\) broadcasts its opinion $\hat{z}_i(t)$ along with its \textit{id} at a regular interval determined by $\Delta t_{\mathrm{tx}}$ and maintains a local memory of the estimates received from its peers. The memory is time-bound, with a horizon equal to the update interval~$T_{\mathrm{up}}$. $M_i(t)$ denotes the set of peers whose latest estimate is available to robot~$i$ at time~$t$. For each peer $j\in M_i(t)$, robot $i$ stores only the most recently received estimate $\hat{z}_j(t_{ij})$, where $t_{ij}$ denotes the corresponding reception time. Upon receiving a new estimate from the peer~$j$, the stored value is replaced and $t_{ij}$ is updated. An estimate is removed from memory when $t-t_{ij}>T_{\mathrm{up}}$. The social information at time~$t$ is defined by
\begin{equation}
    \overline{s}_i(t)=\frac{1}{|M_i(t)|}\sum_{j\in M_i(t)}\hat{z}_j(t_{ij}).
    \label{eq:social_info_dgf}
\end{equation}

The estimate of robot $i$ is updated as
\begin{equation}
    \hat{z}_i(t) = \omega_{ii}(k)\cdot z_i(t) + (1 - \omega_{ii}(k)) \cdot \overline{s}_i(t).
    \label{eq:opinion_update}
\end{equation}

%\paragraph{Accumulating influence and adapting self-confidence}

The self-confidence $\omega_{ii}$ of each robot is adapted according to its observed influence on the estimates of its peers. Influence is inferred from changes between consecutive estimates received from the same peer. Let $\hat{z}_i(t_b)$ denote the most recently broadcast estimate of robot~$i$. Suppose robot~$i$ stores an estimate $\hat{z}^{\mathrm{old}}_j$ from peer~$j$ and subsequently receives an updated estimate $\hat{z}^{\mathrm{new}}_j$. Robot~$i$ measures the distance of both estimates from its own broadcast reference by
\begin{equation}
    d_i^{new} = \left|| \hat{z}^{\mathrm{new}}_j-\hat{z}_i(t_b)|\right|_2, \quad
    d_i^{old} = \left|| \hat{z}^{\mathrm{old}}_j-\hat{z}_i(t_b)|\right|_2.
\end{equation}
A decrease in $d_i^{new}$ relative to $d_i^{old}$ indicates that peer~$j$'s estimate has moved closer to the broadcast estimate of robot~$i$ and vice versa. The corresponding influence contribution is therefore given by 
\begin{equation}
    I_j=\max\left(0, d^{\mathrm{old}}_{ij}-d^{\mathrm{new}}_{ij}\right). 
\end{equation}
If no previous estimate from peer~$j$ is available in memory, the newly received estimate is stored without computing an influence contribution. For each repeated message received from any peer~$j$ during an update interval~$k$, the corresponding influence contribution is added to robot~$i$'s influence accumulator~$I_i(k)$. The total influence accumulated by robot~$i$ over the update interval is thus given by 
\begin{equation}
    I_{i}(k)=\sum_{r \in \mathcal{R}_i^{k}} I_{j}
\end{equation}
where $\mathcal{R}_i^{k}$ denotes the set of repeated-message events observed by robot $i$ during update interval $k$. For the first update interval, when no influence history is available, the reference value is initialised as $I_i(0)=0$.

At the end of an update interval, the accumulated influence is compared with the mean of the robot's recent historical influence. Let
\begin{equation}
    I_i^H=\frac{1}{|H_i|}\sum_{h=1}^{H_i}I_i{(k-h)}
\end{equation}
denote the mean influence accumulated over the preceding update intervals, where $H_i=\{h \in \{1,\ldots,\min(H,k)\}\mid I_i(k-h)>I_{\mathrm{thresh}}\}$ with a maximum history length of $H$ and minimum influence threshold $I_\mathrm{thresh}$. The change in self-confidence is then computed from the difference between the current accumulated influence $I_i(k)$ and this recent historical reference $I_i^H$ as
\begin{equation}
    \Delta\omega_{ii}(k)=\operatorname{clip} \left[\eta \cdot \left(I_i(k)-I_i^{H}\right), -\Delta_{\max}, \Delta_{\max}\right]
\end{equation}
where $\eta$ is the influence learning rate and $\Delta_{\max}$ limits the maximum confidence change between consecutive update intervals. The influence accumulator is reset (i.e., $I_{i}(k)=0$) at the end of each update interval $k$. As $I_i^{H}$ is the reference influence level, a finite history
allows it to adapt over time and prevents observations from the distant past from affecting the self-confidence dynamics. This is particularly important when the sensing quality of individual robots changes during operation. Once the opinions of robot~$i$ and its peers become sufficiently similar (i.e., a consensus is reached), repeated messages produce little or no additional influence, causing $I_i(k)$ to approach zero. Including such update intervals would progressively reduce the historical mean $I_i^{H}$, even though this reduction is a consequence of consensus rather than a change in the robot's relative influence. Therefore, we include only update intervals satisfying $I_i(k)>I_{\mathrm{thresh}}$ in the historical reference where $I_{\mathrm{thresh}}>0$.

The self-confidence for the subsequent update interval is updated according to
\begin{equation}
    \omega_{ii}(k+1)= \operatorname{clip} [\omega_{ii}(k)+\Delta\omega_{ii}(k), \omega_{min}, 1.0-\omega_{min}].
    \label{eq:self_conf_update}
\end{equation}
A robot increases its self-confidence when its accumulated influence during the current update interval exceeds its recent historical level, whereas below-average influence results decrease self-confidence.

\subsubsection{Asymmetric Peer Weighting}

We propose DGF-Asym, an extension of the DeGroot-Friedkin model with asymmetric peer weighting; that is, we change the way in which social information~$\overline{s}_i$ is constructed. 
In the standard DGF model, the relative interpersonal weights of peers remain fixed. Here, we replace this fixed weighting with a peer-specific weighted averaging, allowing robot~$i$ to assign different levels of influence to different neighbors when forming its social estimate. In this way, peers with potentially more accurate estimates gain influence, while those with less reliable estimates, for example, due to noisy or faulty sensors, have less influence on the overall estimate.

We introduce a retrospective predictive-performance measure to distinguish between peers of differing information quality. Each robot~$i$ maintains a short-term history of $\lambda=10\,s$ of its own estimates and of the estimates received from its peers.
Consider the current estimate $\hat{z}_i(t)$ of robot $i$, its own past estimate $\hat{z}_i(t-\lambda)$, and the past estimate $\hat{z}_j(t-\lambda)$ received from peer $j$. The usefulness of peer $j$ from the perspective of robot $i$ is defined as
\begin{equation}
    U_{ij}(t-\lambda)= \left||\hat{z}_i(t-\lambda) - \hat{z}_i(t)|\right|_2 - \left||\hat{z}_j(t-\lambda) - \hat{z}_i(t)|\right|_2.
\end{equation}

The first term measures the deviation of robot $i$'s own previous estimate from its current estimate, while the second term measures the deviation of peer $j$'s previous estimate from robot~$i$'s current estimate. Hence
\begin{itemize}
    \item $U_{ij}(t-\lambda)>0$ indicates that the past estimate of peer $j$ was closer to the estimate eventually reached by robot~$i$ compared to robot $i$'s own past estimate, so $j$ is considered useful.
    \item $U_{ij}(t-\lambda)<0$ indicates that robot $i$'s own past estimate was closer to its current estimate than the corresponding estimate from peer $j$, so $j$ is considered less useful.
    \item $U_{ij}(t-\lambda)\approx 0$ indicates that the past estimate of~$j$ provided not much information about its usefulness (approximately equal predictive proximity).
\end{itemize}
Newly encountered peers are initialized with $U_{ij}=1$.

Robot~$i$ maps the usefulness $U_{ij}(t-\lambda)$ to a non-negative peer weight $\omega_{ij}(t)$ through the sigmoid function
\begin{equation}
    \omega_{ij}(t) = \frac{10}{1+9\cdot e^{-U_{ij}(t-\lambda)}}.
\end{equation}
This mapping assigns neutral weight \(\omega_{ij}=1\) when \(U_{ij}=0\). Positive usefulness increases the weight monotonically toward an upper limit of \(10\), whereas negative usefulness decreases it toward \(0\).

In our asymmetric peer weighting formulation, Eq.~\ref{eq:social_info_dgf} is updated to the normalized weighted average 
\begin{equation}
    \overline{s}_i = \frac{\sum_{j\in M_i(t)}\omega_{ij}(t)\cdot \hat{z}_j(t_{ij})}
    {\sum_{j\in M_i(t)} \omega_{ij}(t)}.
    \label{eq:social_info_asym_dgf}
\end{equation}
Peers assessed as more useful contribute more strongly to the social information of robot~$i$ than less useful peers. The subsequent opinion update remains as in Eq.~\ref{eq:opinion_update}. The self-confidence $\omega_{ii}(k)$ is updated at the end of each update interval according to Eq.~\ref{eq:self_conf_update}. In contrast, the peer weight $\omega_{ij}(t)$ is updated whenever a new opinion is received from peer~$j$, provided that a previous opinion from $j$ is available within the short-term memory at $(t-\lambda)\,s$.

\section{EXPERIMENTAL EVALUATION}

\subsection{Setup}

We evaluate our method using a collective navigation task in the physics-based multi-robot simulator ARGoS \cite{Pinciroli:SI2012}. A team of $N$~TurtleBot~4 robots starts from random poses in an obstacle-free $50\,\text{m}\times50\,\text{m}$ arena and has to navigate toward a common time-varying goal~$p^{gt}(t)=[x_{p^{gt}}(t), y_{p^{gt}}(t)]^\top$.
Each robot receives a noisy pose estimate 
\(
    \hat{p}_i(t) = p_i(t)+\nu_{i,p}(t),
    \hat{\psi}_i(t) = \psi_i(t)+\nu_{i,\psi}(t),
\)
with ground-truth position and heading~$p_i(t)=[x_i(t),y_i(t)]^\top$ and $\psi_i(t)$, $\nu_{i,p}(t) \sim \mathcal{N}\!\left(0, \sigma_p^2 \right)$, $\nu_{i,\psi}(t)\sim\mathcal{N}(0,\sigma_\psi^2)$, $\sigma_p^2=0.05\,\mathrm{m}^2$, and $\sigma_\psi^2=0.05\,\text{rad}^2$. The parameters are set to $\omega_{ii}(0)=0.5$, $\omega_{min}=0.005$, $\eta=0.01$ and $\Delta_{max}=0.1$.

When the goal is visible, robot~\(i\) obtains a noisy camera observation in its local frame at every time step~\(t\), given by
\begin{equation}
\hat{p}_i^g(t)=r_i(t)
\begin{bmatrix}
\cos\theta_i(t)\\
\sin\theta_i(t)
\end{bmatrix}
+r_i(t)\epsilon_i(t)
\begin{bmatrix}
\cos\phi_i(t)\\
\sin\phi_i(t)
\end{bmatrix},
\end{equation}
where $r_i(t)$ and $\theta_i(t)$ are the true goal range and bearing, $\epsilon_i(t)\sim\mathcal{N}(0,\sigma_s^2)$, and $\phi_i(t)\sim\mathcal{U}(0,2\pi)$. The observation noise therefore scales linearly with distance. Using its estimated pose, the robot transforms this observation into global coordinates: 
\begin{equation}
    z_i(t) = \hat{p}_i(t)+R\!\left(\hat{\psi}_i(t)\right)\hat{p}_i^g(t)
\end{equation}
where \(R(\psi)\)~is the planar rotation matrix. The observation is fused with the social information to obtain the goal-position opinion~$\hat{z}_i(t)$ (see Eqs.~\ref{eq:social_info_dgf},~\ref{eq:social_info_asym_dgf}). Robots have no prior knowledge of their sensor reliability and broadcast their opinions at robot-specific intervals~$\Delta t_i^{\mathrm{tx}} \sim \mathcal{U}(3,5)\,s$, sampled at initialization. 

Each robot drives toward the goal following the vector from its current position~$\hat{p}_i(t)$ to its current estimated goal~\(\hat{z}_i(t)\). Reactive IR-sensor-based collision avoidance prevents robot-robot and robot-wall collisions. Robots stop within \(1\,\mathrm{m}\) of their current goal opinion and also will not execute collision avoidance to limit unnecessary motion near the estimated goal.

Each $2700\,s$ run comprises an initial stationary goal period of~$500\,s$ followed by four cycles of $50\,s$ goal relocation and a $500\,s$ evaluation episodes~$\ell\in\mathcal{L}$ with a stationary goal. We conduct ten independent runs per experimental condition, yielding $|\mathcal{L}|=40$~evaluation episodes per configuration.

To test robustness, a fraction $\rho = N_f/N$ with number of defective robots~$N_f$ produces uninformative goal observations~$z_i(t)\sim\mathcal{U}([0,50]^2)$.
The algorithm is not informed of the faults or affected robot identities. 
The same model applies when occlusion or camera orientation prevents goal observation, that is, each location is considered equally likely (i.e., a maximum-ignorance model). 

Defective identities change at fixed intervals while~\(\rho\) remains constant, modeling alternating sensor failures and recoveries. Nominal experiments use \(\rho=0\), with the same multi-robot system size and parameters, to establish a baseline for performance comparison. 

\subsection{Metrics}

\label{sec:metrics}
We evaluate our method and the baselines in terms of decision accuracy, stability, and success rate. The individual opinion error of robot \(i\) at time $t$ is defined as
\begin{equation}
E_i(t)=\left|\hat{z}_i(t)-p^{gt}(t)\right|_2, 
\label{eq:individual_opinion_error}
\end{equation}
and the collective opinion error at time $t$ is defined as 
\begin{equation}
 E_c(t)=\left|\frac{1}{N}\sum_{i=1}^N\hat{z}_i(t) - p^{gt}(t)\right|_2.
\end{equation}

For each evaluation episode \(\ell\), \(t_{\ell}\in[0,T_{\ell}]\,s\) denotes the elapsed time within that episode. Let \(E_c^{(\ell)}(t_{\ell})\) denote the collective opinion error in evaluation episode \(\ell\) at relative time \(t_\ell\). The median collective opinion error $E^\text{med}_c(t_{\ell})$ is the median of \(E_c^{(\ell)}(t_{\ell})\) over all evaluation episodes $\ell \in \mathcal{L}$ at each relative time $t_{\ell}$.

From the nominal experiments, the median collective opinion error $E^\text{med}_c(t_{\ell})$ is obtained. The nominal error threshold $E_c^{\mathrm{nom}}$ is the median of $E^{med}_c(t_{\ell})$ over time $t_{\ell} \in [0,T_{\ell}]$.

An evaluation episode~\(\ell\) is considered successful if there exists a continuous interval of time \(\tau\) during which the collective opinion error \(E_c^{(\ell)}\) remains below the nominal threshold $E_c^{\mathrm{nom}}$. The success rate for a given minimum in-threshold duration \(\tau\) is defined as $SR_{\tau} = |\mathcal{L}_s^{\tau}|/|\mathcal{L}|$, representing the fraction of successful episodes, with the set of successful evaluation episodes~$\mathcal{L}_s^{\tau}$ for a given $\tau$. Response time for each successful episode $\ell \in \mathcal{L}_s^{\tau}$ is defined as
\begin{equation}
    t_{(\ell, \tau)}^{dec} = \operatorname{min}\left\{t_{\ell}\;\middle|\; [0,T_{\ell}-\tau] \subseteq \left\{t_{\ell}:E_c^{(\ell)}(t_{\ell})\leq E_c^{\mathrm{nom}} \right\}\right\}.
\end{equation}
$T^{dec}_{\tau} = \operatorname{median}_{\forall \ell \in \mathcal{L}_s^{\tau}}\{t_{(\ell, \tau)}^{dec}\}$ is the median time to satisfy the success condition across all successful evaluation episodes.

To evaluate the final decision accuracy of the collective, we average the collective opinion error over the final self-confidence interval~\(k\), which has a duration of ~$T_{\mathrm{up}}$, of each evaluation episode. The mean collective opinion error is given as
\begin{equation}
\bar{E}_c^{(\ell)} = \frac{1}{T_{\mathrm{up}}} \sum_{t_{s}=T_{\ell}-T_{\mathrm{up}}}^{T_{\ell}} E_c^{(\ell)}(t_{s}).
\end{equation}
We report $E^{con}_c$, that is, the median of \(\bar{E}_c^{(\ell)}\) across all episodes \(\ell\in\mathcal{L}\), including unsuccessful ones, to avoid bias toward better-performing trials.

The mean robot opinion error during the final update interval is given as
\begin{equation}
    \bar{E}_r^{(\ell)}=\frac{1}{T_{\mathrm{up}}} \sum_{t_{s}=T_{\ell}-T_{\mathrm{up}}}^{T_{\ell}} \bar{E}^{(\ell)}_r(t_{s})
\end{equation}
where $\bar{E}_r^{(\ell)}(t_{s})$ is the mean opinion over the group size $N$ at time $t_s$. The robot opinion error at consensus $\bar{E}_r^{con}$ is the median of $\bar{E}_r^{(\ell)}$ over all episodes $\ell \in \mathcal{L}$.

The temporal stability of the individual and collective opinions is also evaluated over the final update interval $t_s \in [T_{\ell}-T_\mathrm{up}, T_{\ell}]$. For each robot $i$, we first calculate the mean robot opinion $\bar{z}_i^{con}$ during the final update interval for an episode $\ell$, that is,
\begin{equation}
    \bar{z}_{(i,\ell)}^{con} = \frac{1}{T_\mathrm{up}}\sum_{t_{s}=T_{\ell}-T_{\mathrm{up}}}^{T_{\ell}} \hat{z}_{(i,\ell)}(t_s),
\end{equation}
where $\hat{z}_{i,\ell}(t_s)$ is the opinion of robot $i$ at time $t_s$ of episode~$\ell$. The individual robot opinion jitter for an episode is 
\begin{equation}
    J_r^{(\ell)} = \sqrt{\frac{1}{N}\frac{1}{T_\mathrm{up}}\sum_{i=1}^N\sum_{t_{s}=T_{\ell}-T_{\mathrm{up}}}^{T_{\ell}} \left ||\bar{z}_{(i,\ell)}^{con} - \hat{z}_{(i,\ell)}(t_s)|\right|_2^2}\,,
\end{equation}
and robot opinion jitter $J_r$ is the median of $J_r^{(\ell)}$ over all episodes. The collective opinion error jitter for episode $\ell$ is given by 
\begin{equation}
J_c^{(\ell)}=\sqrt{\frac{1}{T_{\mathrm{up}}}
\sum_{t_s=T_\ell-T_{\mathrm{up}}}^{T_\ell}
\left(
E_c^{(\ell)}(t_s)-\bar{E}_c^{(\ell)}
\right)^2
}
\end{equation}
and collective opinion error jitter $J_c$ is the median of $J_c^{(\ell)}$ over all episodes. A lower jitter indicates a more stable opinion.

\subsection{Baselines}
We compare our approach with AOL-C and AOL-1p, two variants of the Autonomous Online Learning framework proposed by Gupta et al.~\cite{Gupta2024AOL}. These decentralized methods learn fusion weights online without ground-truth feedback. Developed for cooperative target monitoring under noise, sensor failures, and intermittent visibility, they provide relevant baselines for adaptive fusion under unreliable sensing.

Both methods combine local and social estimation through exchange of intermediate estimates and their weights. AOL-C compares the current observation and previous estimate with the locally available intermediate estimate having the highest weight. AOL-1P uses a perturbation-greedy reward to explore alternative weightings and reinforce changes that improve target-detection confidence. We adapt both methods to our sensing model: loss of visibility produces a maximum-ignorance observation rather than retention of the previous estimate. We also adjust AOL parameters to accommodate sensing without a finite range limit, where goal visibility depends on occlusion and the robot's field of view. 

\subsection{Results}

\begin{table*}[!ht]
    \centering
    \vspace{5.1pt}
    \caption{Collective decision performance of different methods. Tested under $N=5$, $\rho=0.2$, $\sigma_s^2=0.05$, $I_{thresh}=1$. \\ Metrics are reported as
    median [$Q_{25}$--$Q_{75}$].}
    \label{tab:collective_decision_performance}
    \footnotesize
    \setlength{\tabcolsep}{6pt}
    \renewcommand{\arraystretch}{1.12}

    \begin{tabular*}{\textwidth}{
        @{\extracolsep{\fill}}
        lcccc
        @{}
    }
        \toprule
        Metric
        & DGF-Asym
        & DGF
        & AOL-C
        & AOL-1P \\
        \midrule

        $E_c^{con}$ $\downarrow$
        & \textbf{1.16} [0.54, 1.72] & 4.39 [2.08, 6.93] & 4.93 [4.70, 5.39] & 5.26 [4.77, 5.63] \\
        $\bar{E}^{con}_r$ $\downarrow$
        & \textbf{1.16} [0.56, 1.72] & 4.61 [2.08, 7.30] & 4.93 [4.70, 5.39] & 5.26 [4.77, 5.63] \\

        \addlinespace[2pt]

        $J_r$ $\downarrow$
        & \textbf{0.09} [0.06, 0.21] & 2.17 [0.11, 6.23] & 7.98 [7.14, 8.62] & 8.17 [7.48, 8.85] \\
        $J_c$ $\downarrow$
        & \textbf{0.07} [0.04, 0.15] & 0.86 [0.10, 2.22] & 7.44 [6.60, 7.98] & 7.55 [6.95, 8.25] \\

        \addlinespace[2pt]

        $SR_{1}$ (\%) $\uparrow$
        & 85\% (34/40) & 42.5\% (17/40) & \textbf{100\%} (40/40) & \textbf{100\%} (40/40) \\
        $T^{dec}_1$ (s) $\downarrow$
        & 348.6 [262.1, 404.2] & 350.0 [175.8, 394.6] & 147.6 [96.5, 169.0] & \textbf{102.3} [79.8, 180.7] \\

        \addlinespace[2pt]

        $SR_{10}$ (\%) $\uparrow$
        & \textbf{82.5\%} (33/40) & 37.5\% (15/40) & 0\% (0/40) & 0\% (0/40) \\
        $T^{dec}_{10}$ (s) $\downarrow$
        & 375.0 [300.0, 450.0] & 351.1 [202.0, 410.3] & -- & -- \\

        \bottomrule
    \end{tabular*}
\end{table*}
\begin{figure*}[t]
    \centering
    \footnotesize{\textbf{N=5}}\\
    \subfloat[$\rho=0.2$ \label{subfig:persistance_SR_COE:a}]{          
        \includegraphics[width=0.32\textwidth]{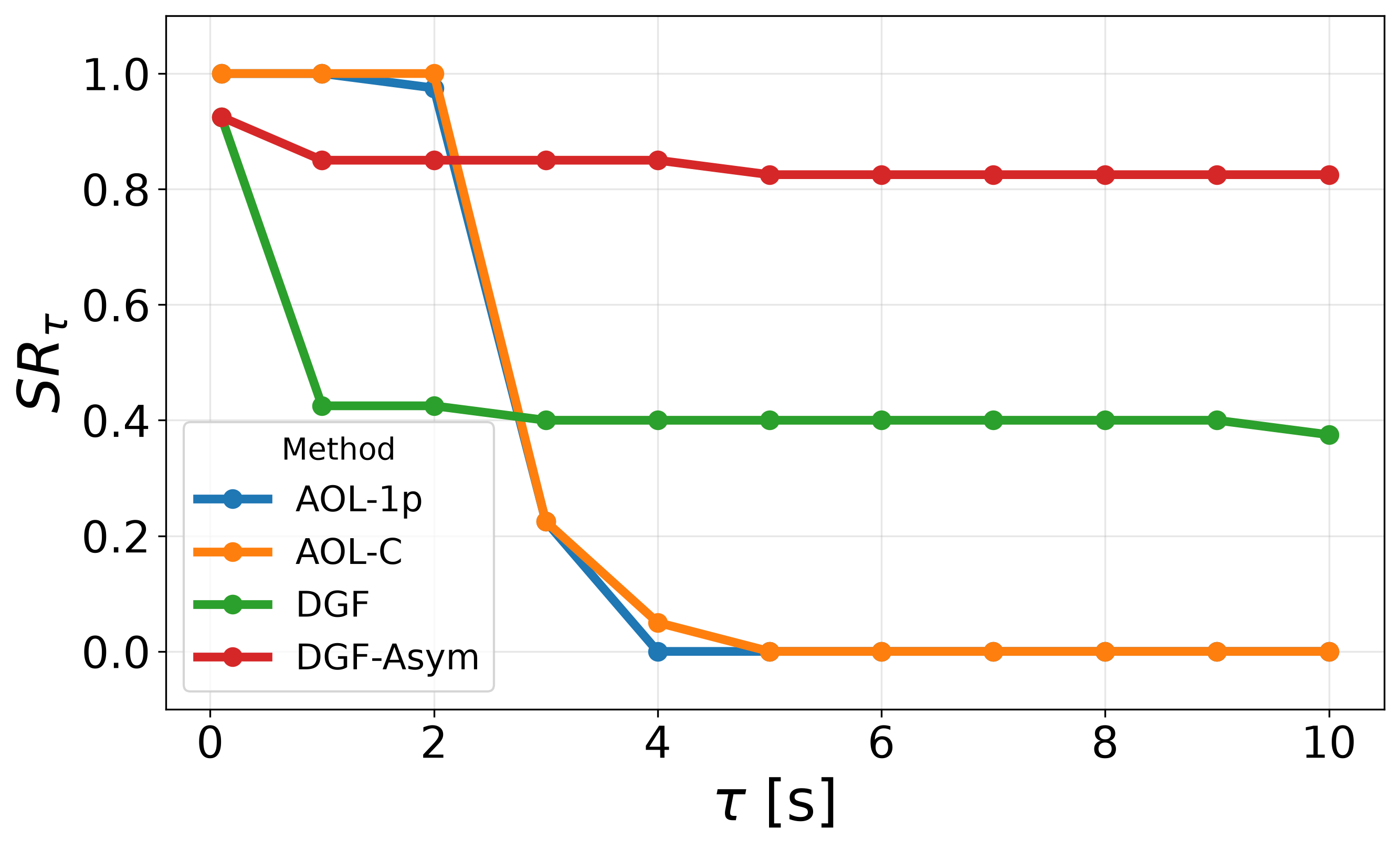}
    }\hfill 
    \subfloat[$\rho=0.4$ \label{subfig:persistance_SR_COE:b}]{       
        \includegraphics[width=0.32\textwidth]{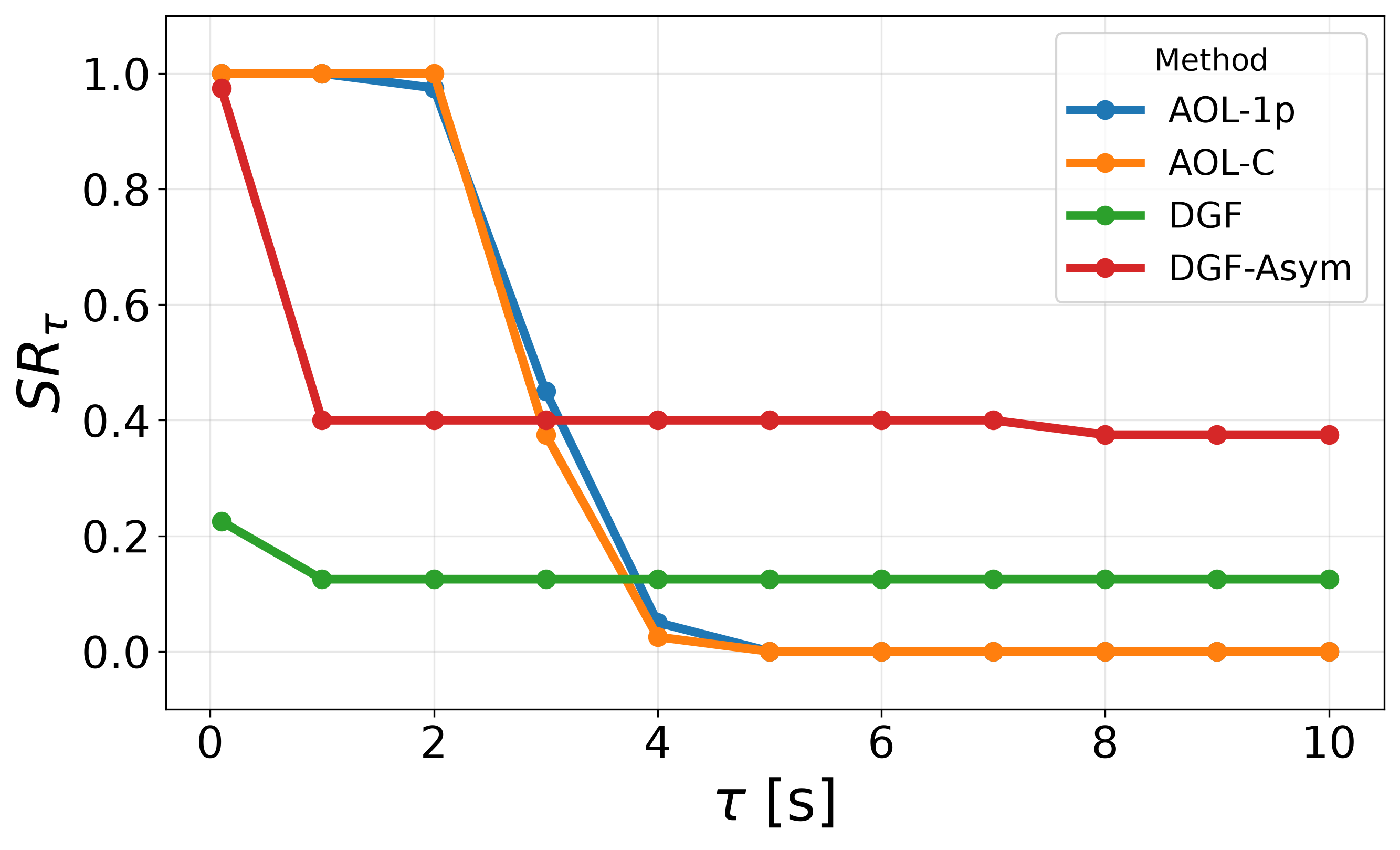}
    }\hfill
    \subfloat[$\rho=0.6$\label{subfig:persistance_SR_COE:c}]{        
        \includegraphics[width=0.32\textwidth]{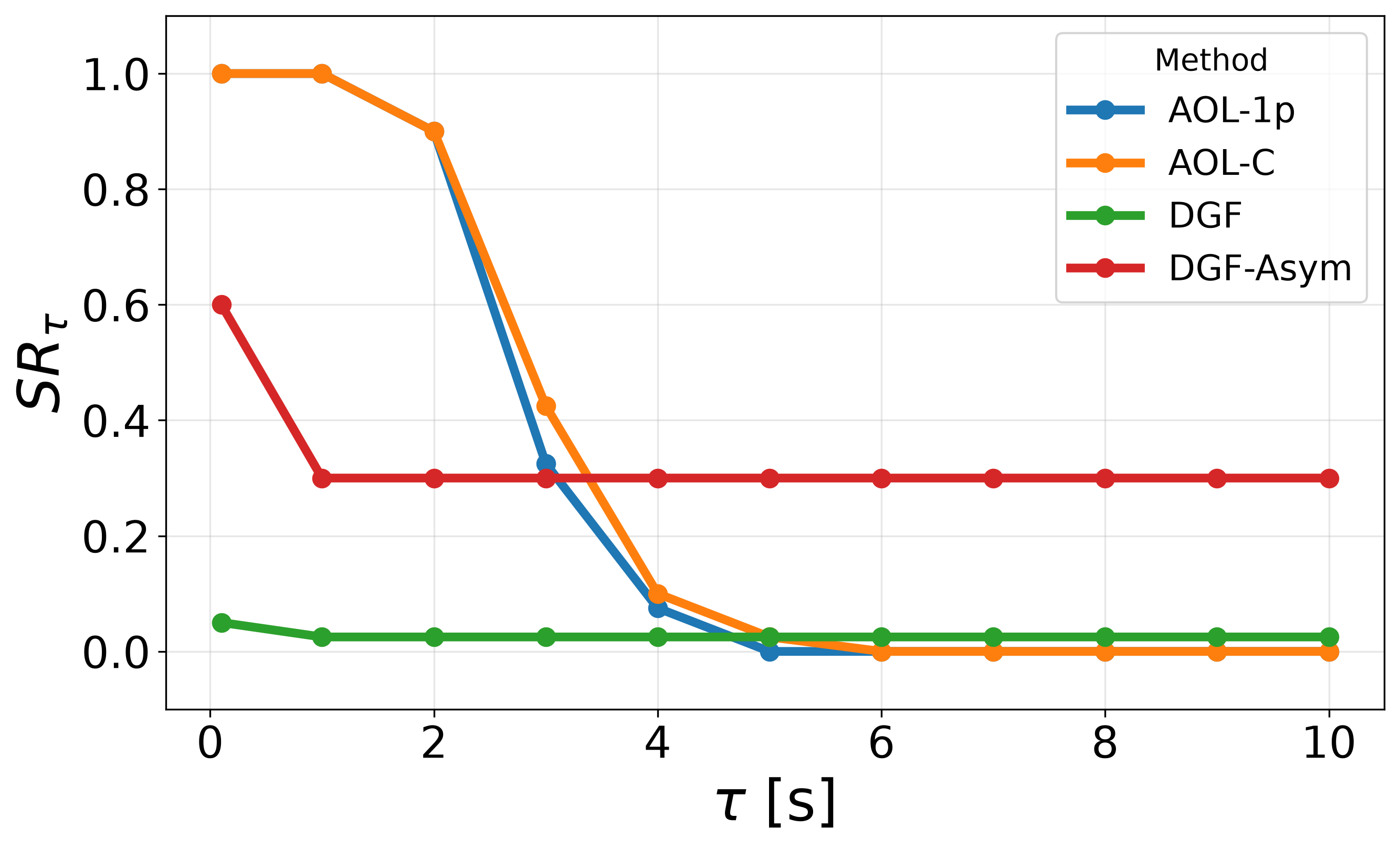}
    }\\[0.5em]
    \footnotesize{\textbf{N=10}}\\
    \subfloat[$\rho=0.2$\label{subfig:persistance_SR_COE:d}]{ 
        \includegraphics[width=0.32\textwidth]{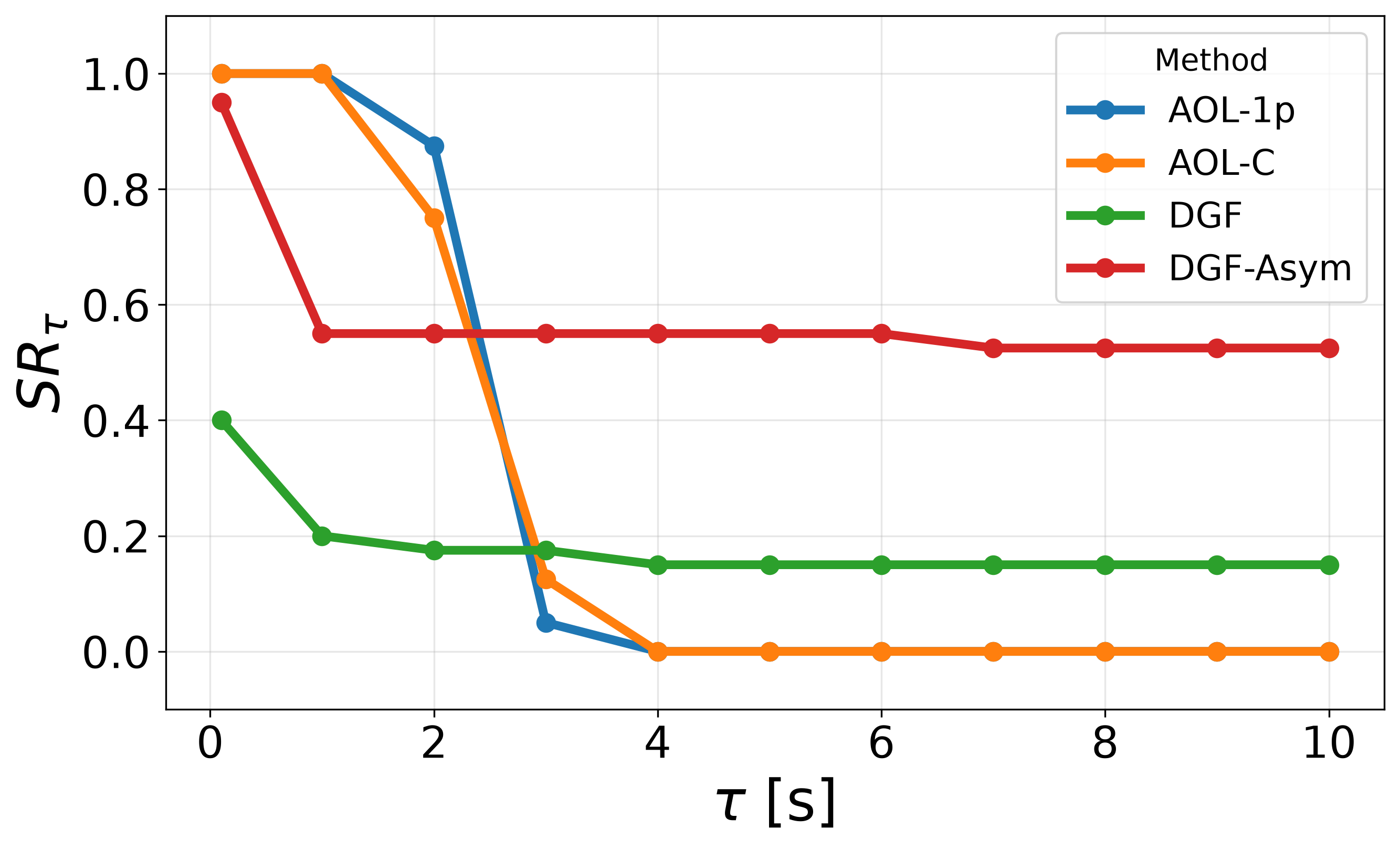}
    }
    \subfloat[$\rho=0.4$\label{subfig:persistance_SR_COE:e}]{
        \includegraphics[width=0.32\textwidth]{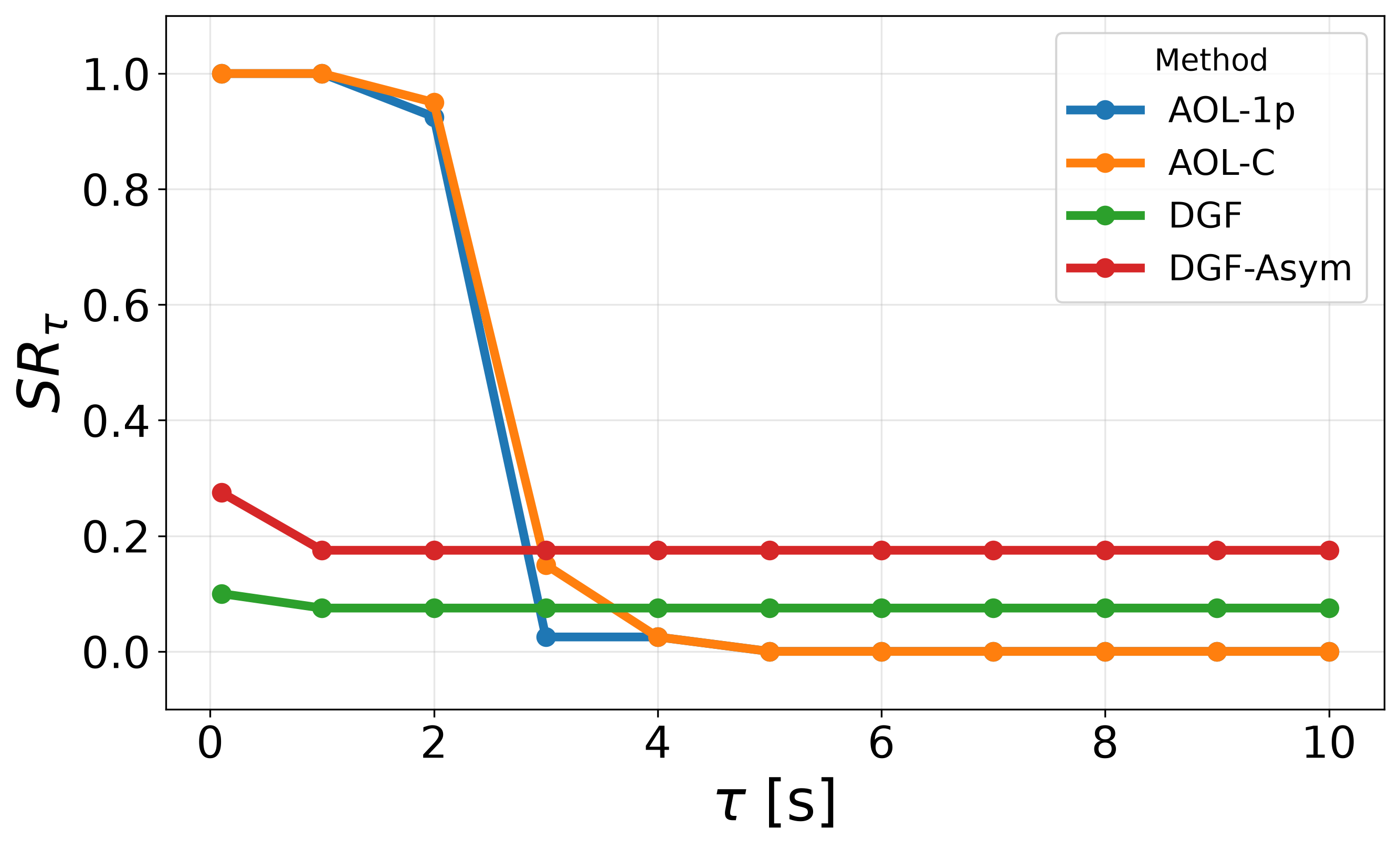}
    }
    \subfloat[$\rho=0.6$ \label{subfig:persistance_SR_COE:f}]{
        \includegraphics[width=0.32\textwidth]{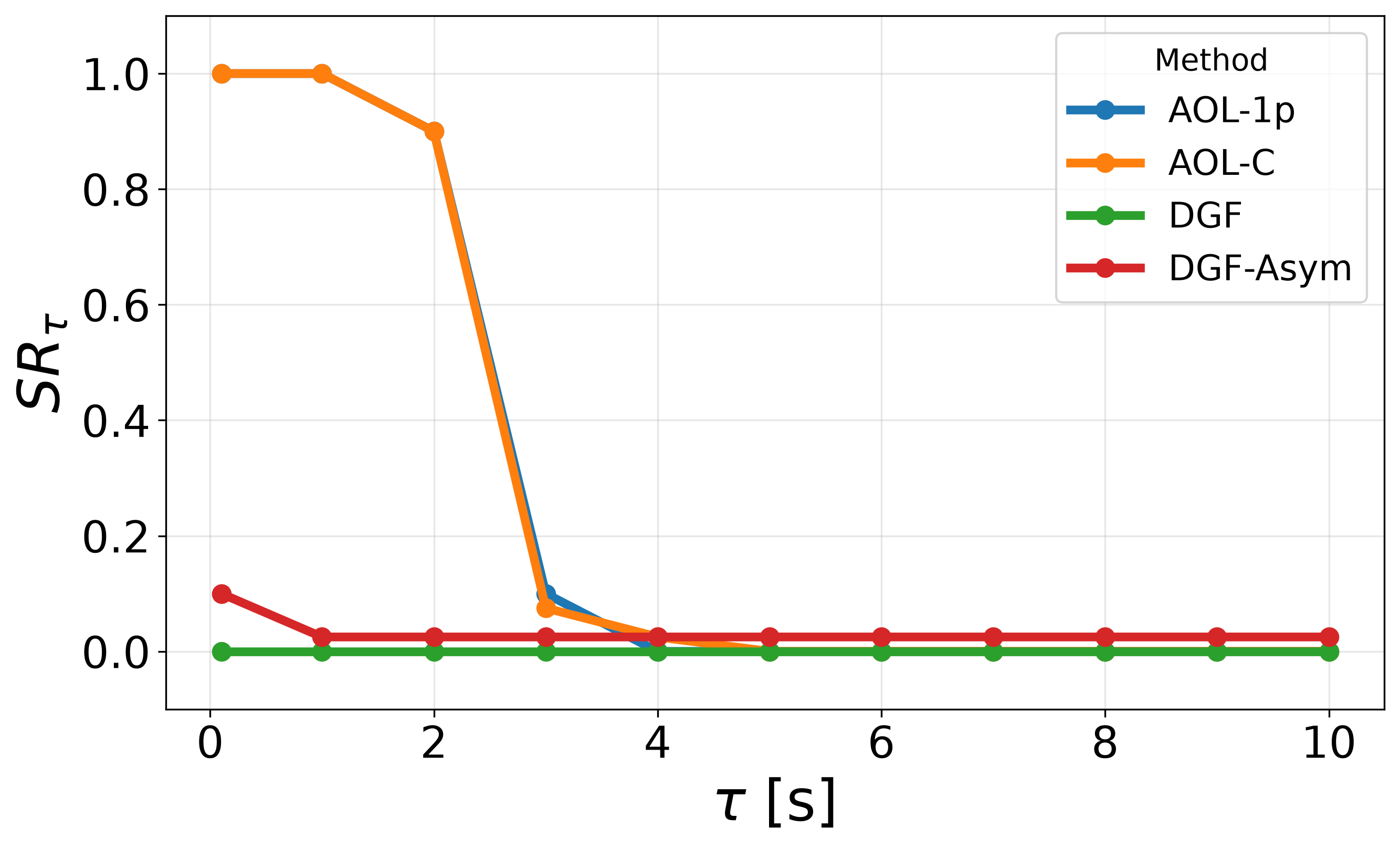}
    }
    \caption{Success rates $SR_{\tau}$ for different combinations of in-threshold durations~\(\tau\), fault ratios~\(\rho\), and multi-robot system sizes~\(N = \{5, 10\}\) with $\sigma^2_s=0.05$ and  $I_{thresh} = 1$.}
    \label{fig:persistance_SR_COE}
\end{figure*}

We first compare the effect of in-threshold durations on the estimation process, testing \(\tau = 1\,\text{s}\) and  \(\tau = 10\,\text{s}\) with \(N=5\) robots, \(\rho=0.2\), and \(I_\text{thresh}=1\). For \(\tau=1\,\text{s}\), AOL-C and AOL-1P both achieve~\(100\,\%\) success, compared with \(85\,\%\) for DGF-Asym, and \(42.5\,\%\) for DGF, see Tab.~\ref{tab:collective_decision_performance}. Increasing the required duration to \(10\,s\) reduces both AOL methods' success rates to \(0\,\%\), while DGF-Asym and DGF retain \(82.5\,\%\) and \(37.5\,\%\), respectively. These results indicate that, although the AOL methods more frequently satisfy the short-duration criterion, DGF-Asym more reliably maintains collective opinions within the threshold over time.

The stability metrics support this distinction. 
DGF-Asym reduces collective opinion error jitter~\(J_c\) by \(91.8\,\%\), \(99.06\,\%\), and \(99.07\,\%\) relative to DGF, AOL-C, and AOL-1P, respectively. The corresponding reductions in median collective opinion error \(E_c\) are \(73.5\%\), \(76.4\%\), and \(77.9\%\). The greater temporal variance of the AOL methods may partly reflect their synchronous communication, which allows noisy observations to propagate across communicating robots within a single update step. DGF-Asym reduces the weight of personal information as consensus emerges and assigns less influence to unreliable peers, limiting sensitivity to noisy inputs and supporting a more stable consensus.

The AOL methods reach the \(\tau=1\,s\) decision condition more quickly in successful episodes; AOL-1P's median decision time~\(T_1^{\mathrm{dec}}\) is \(70.6\%\) shorter than that of DGF-Asym. Together with the success rates at \(\tau = 10\,\text{s}\), this suggests a trade-off: AOL achieves faster short-duration decisions, whereas DGF-Asym more reliably sustains the decision criterion but takes longer to reach it.

\begin{figure*}[t]
    \vspace{5.1pt}
    \centering
        \footnotesize{\textbf{N=5}}\\ \vspace{-0.5em}
        \subfloat[$\sigma^2_s=0.05$, $I_{thresh} = 1$\label{subfig:coe_coeJ_N_5}]{
        \includegraphics[width=0.32\textwidth]{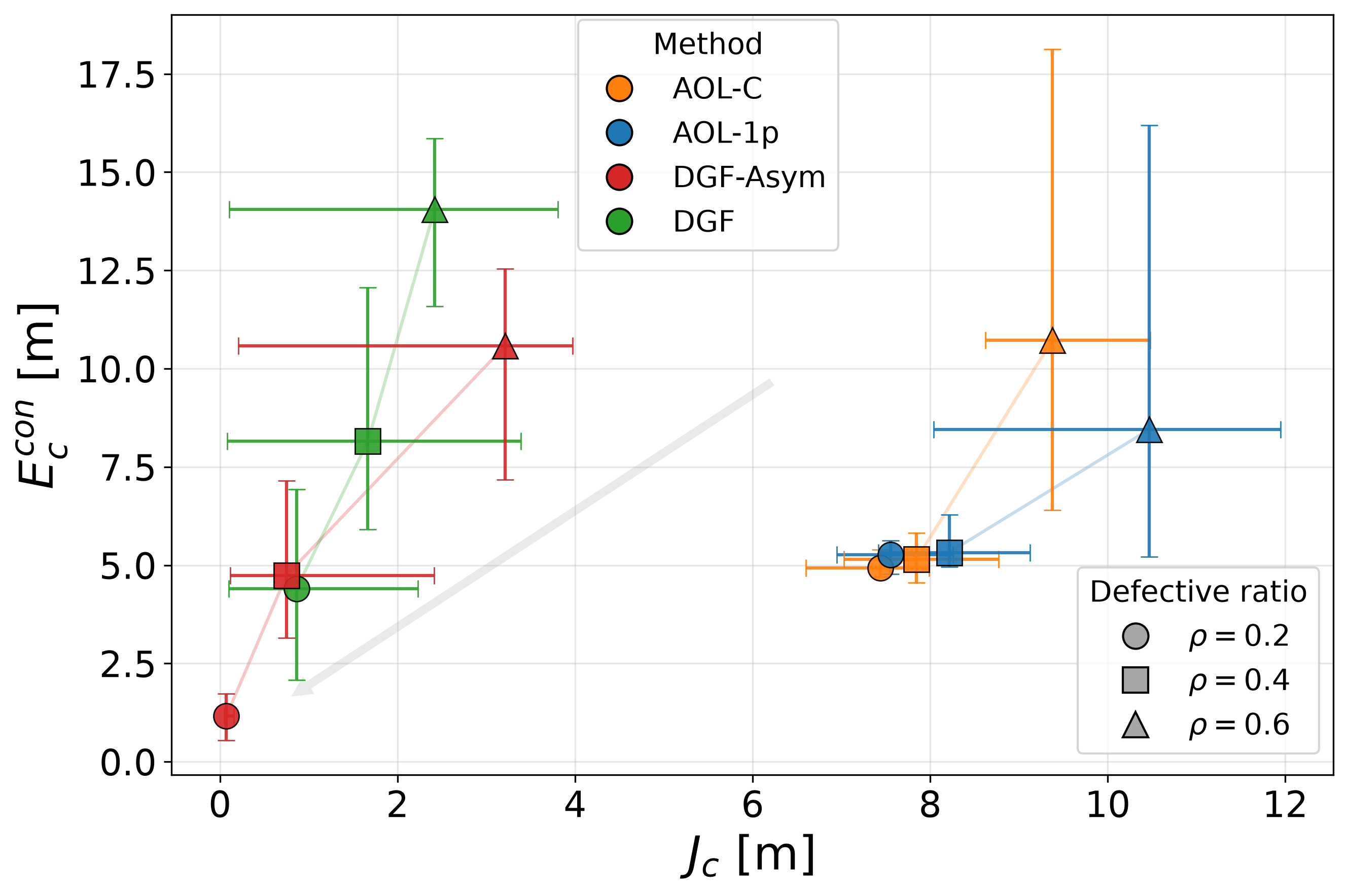}
        } 
    \hfill
    \subfloat[$\rho=0.2$, $I_{thresh} = 1$ \label{subfig:coe_noise_N_5}]{
        \includegraphics[width=0.32\textwidth]{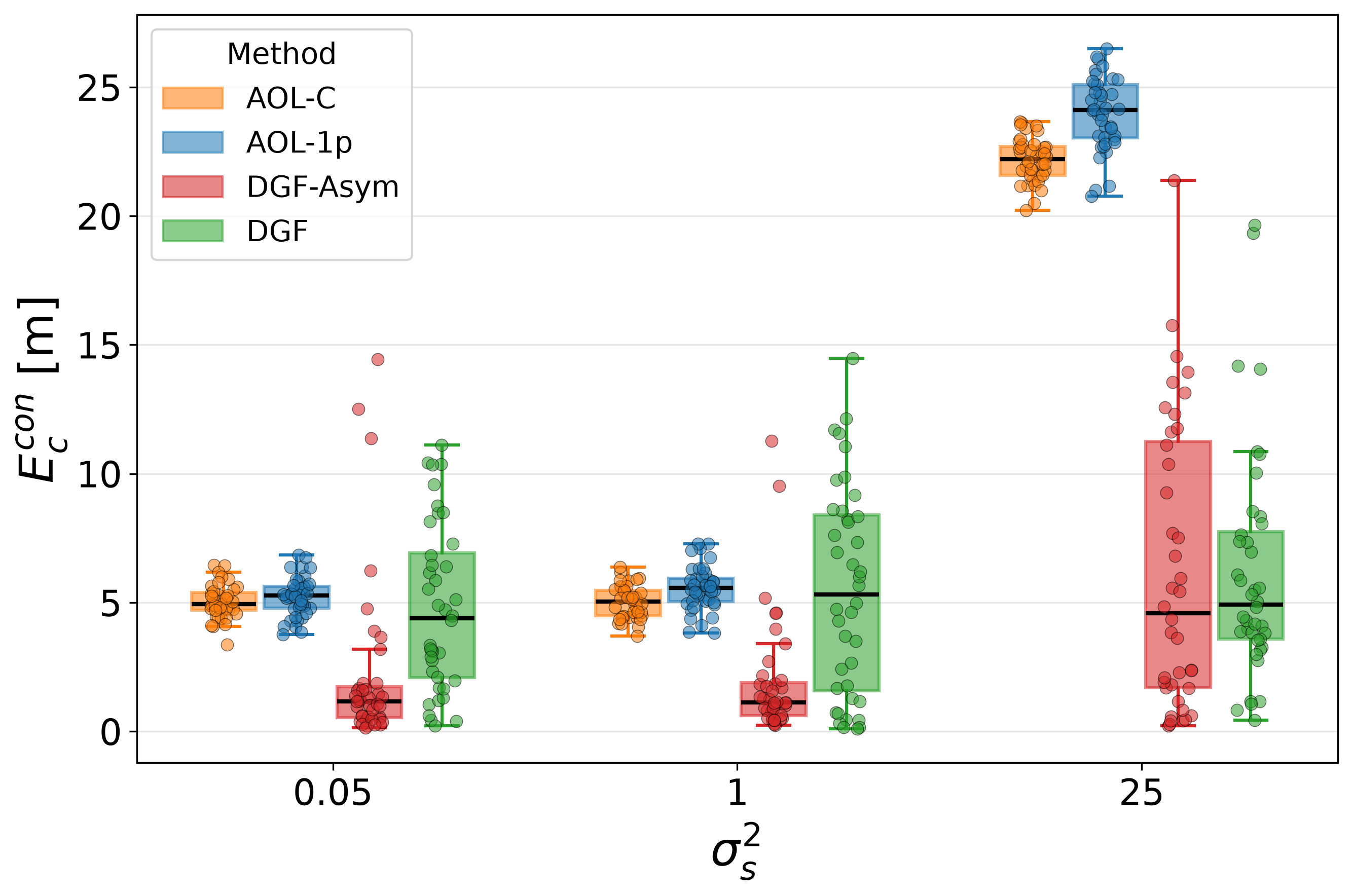}
    }\hfill
    \subfloat[$\rho=0.2$, $\sigma^2_s=0.05$\label{subfig:coe_ithresh_N_5}]{
        \includegraphics[width=0.32\textwidth]{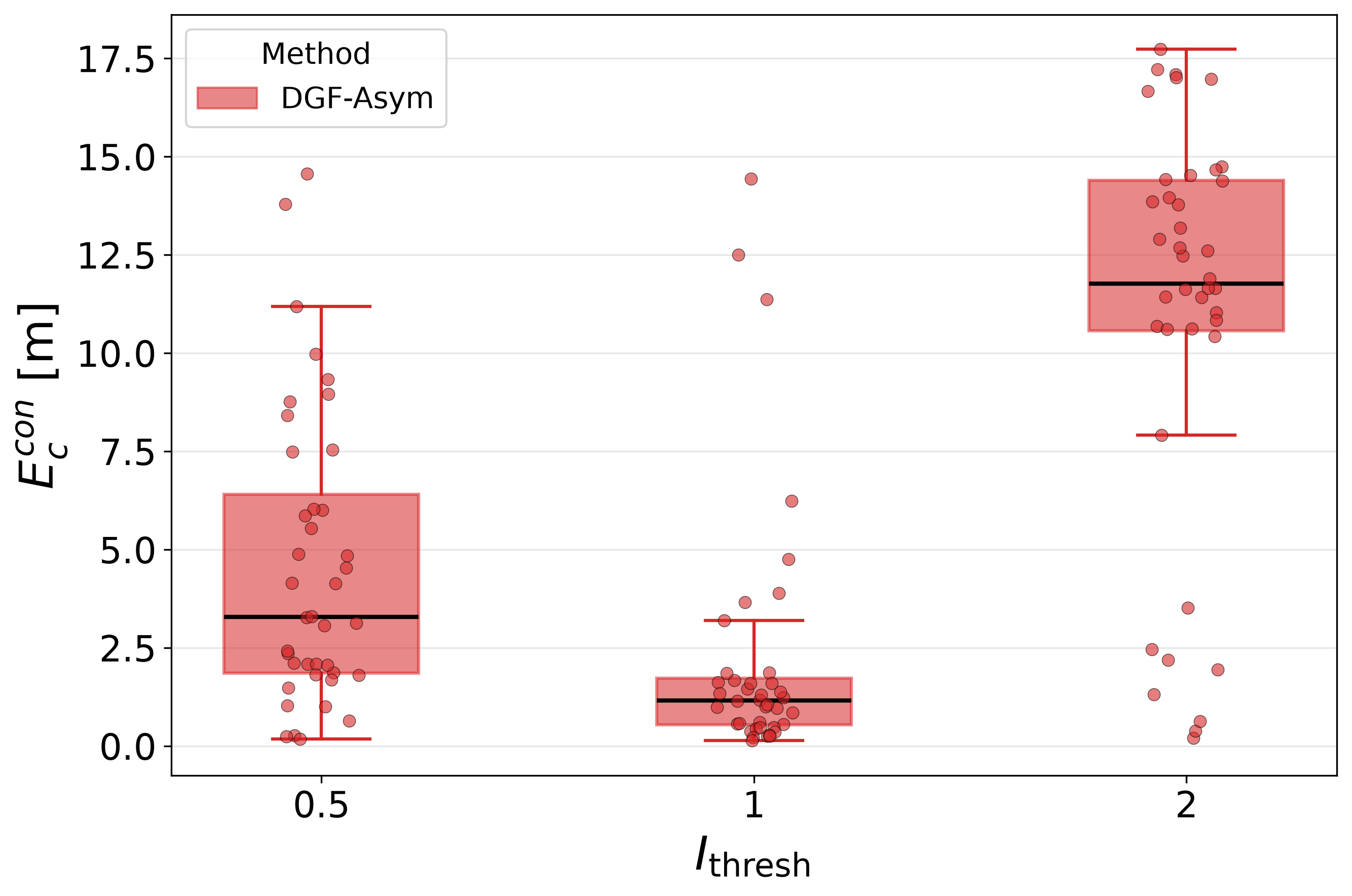}
    }\\[0.5em]
    \footnotesize{\textbf{N=10}}\\ \vspace{-0.5em}
    \subfloat[$\sigma^2_s=0.05$, $I_{thresh} = 1$\label{subfig:coe_coeJ_N_10}]{
            \includegraphics[width=0.32\textwidth]{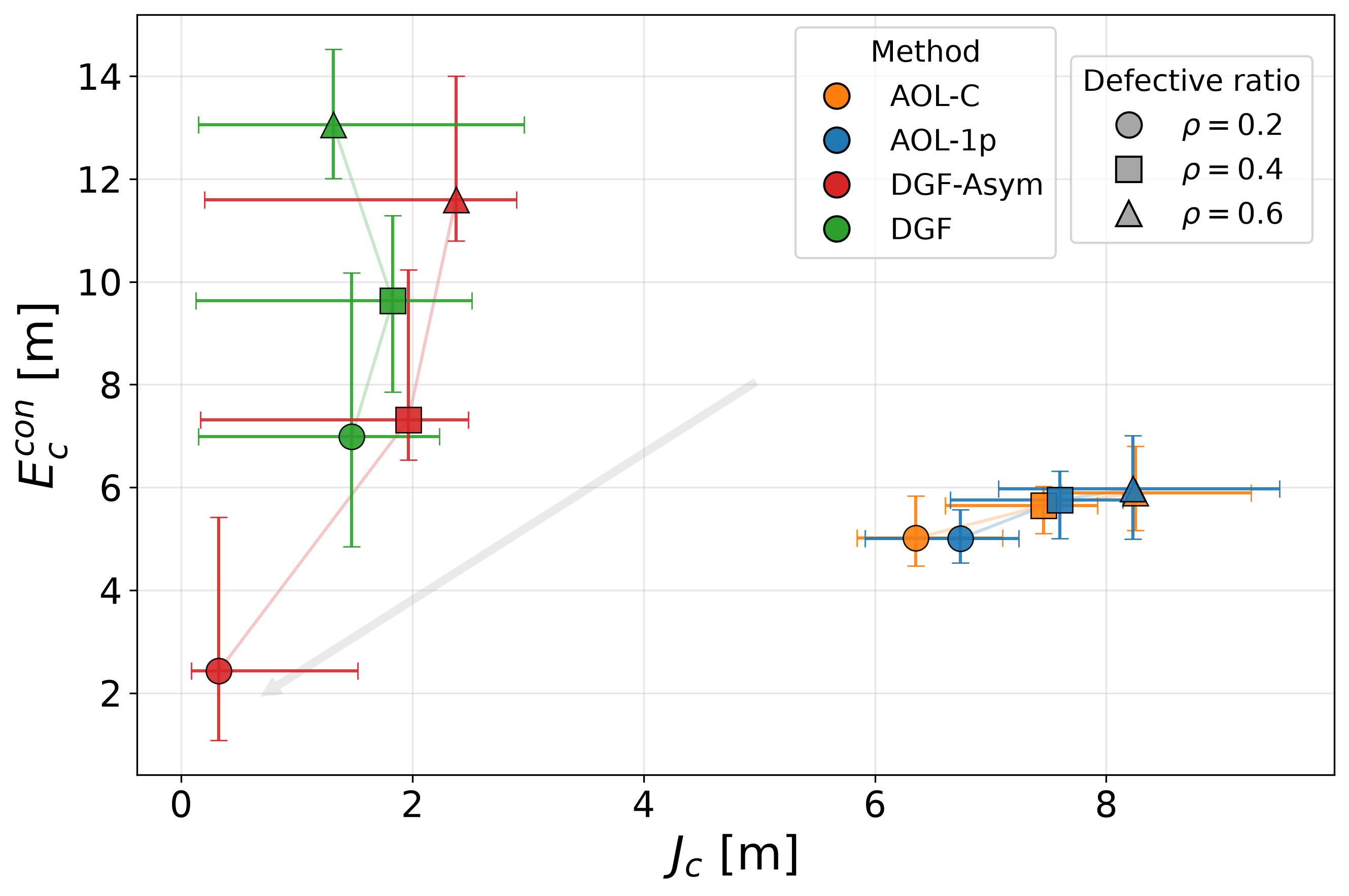}
    }\hfill
    \subfloat[$\rho=0.2$, $I_{thresh} = 1$\label{subfig:coe_noise_N_10}]{\includegraphics[width=0.32\textwidth]{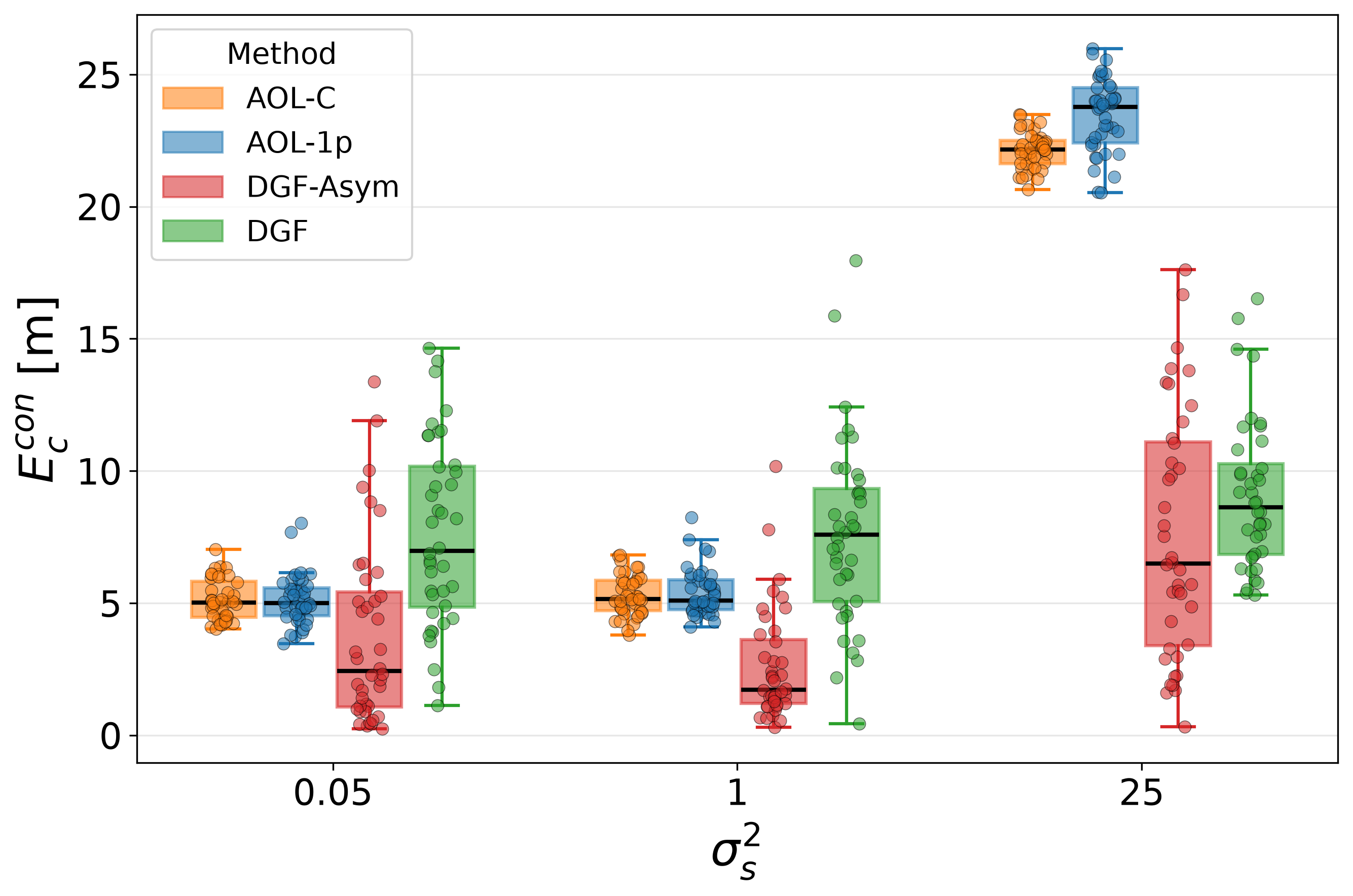}
    }\hfill
    \subfloat[$\rho=0.2$, $\sigma^2_s=0.05$\label{subfig:coe_ithresh_N_10}]{
        \includegraphics[width=0.32\textwidth]{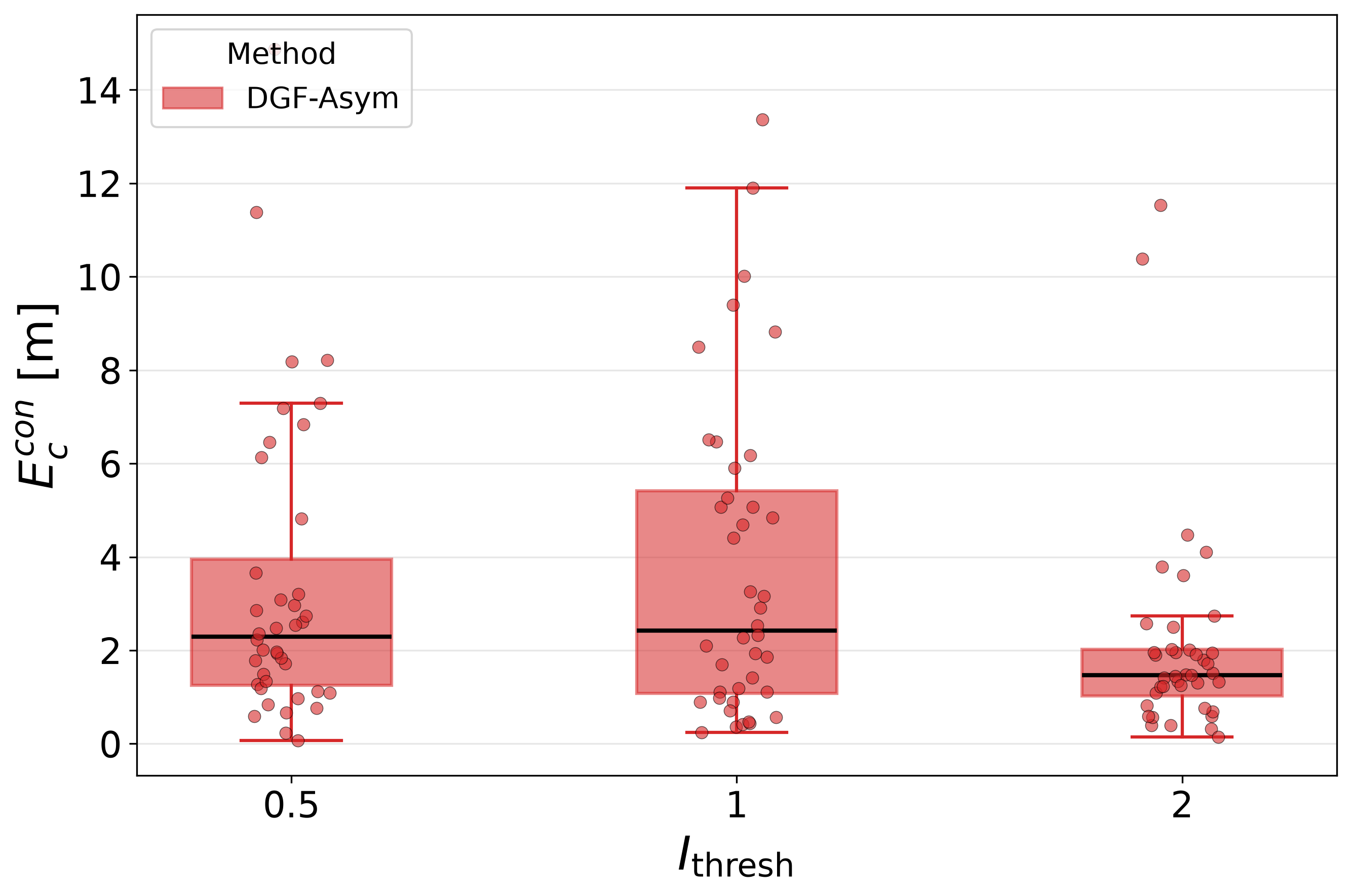}
    }
    \caption{Collective opinion errors $E_c^\mathrm{con}$ along with inter-quartile range for teams of $N=5$ and $N=10$ robots. (a) and (d) show the relationship between collective opinion error \(E_c\) and collective opinion error jitter \(J_c\) for fault ratios \(\rho \in \{0.2,0.4,0.6\}\), with \(\sigma_s^2=0.05\) and \(I_{\mathrm{thresh}}=1\). The lower-left region corresponds to lower error and lower variability. (b) and (e) show collective opinion error across sensor-noise levels \(\sigma_s^2 \in \{0.05,1,25\}\), with \(\rho=0.2\) and \(I_{\mathrm{thresh}}=1\). (c) and (f) show the sensitivity of DGF-Asym's collective opinion error to the influence threshold \(I_{\mathrm{thresh}} \in \{0.5,1,2\}\), with \(\rho=0.2\) and \(\sigma_s^2=0.05\). \textit{Note: The limits of the axes vary between plots; this is done for ease of visualization of close values.}}
    \label{fig:combined_results}
\end{figure*}

\begin{figure}
    \centering
    \subfloat[DGF-Asym\label{subfig:faulty_time_dgfasym}]{
    \includegraphics[width=0.75\linewidth]{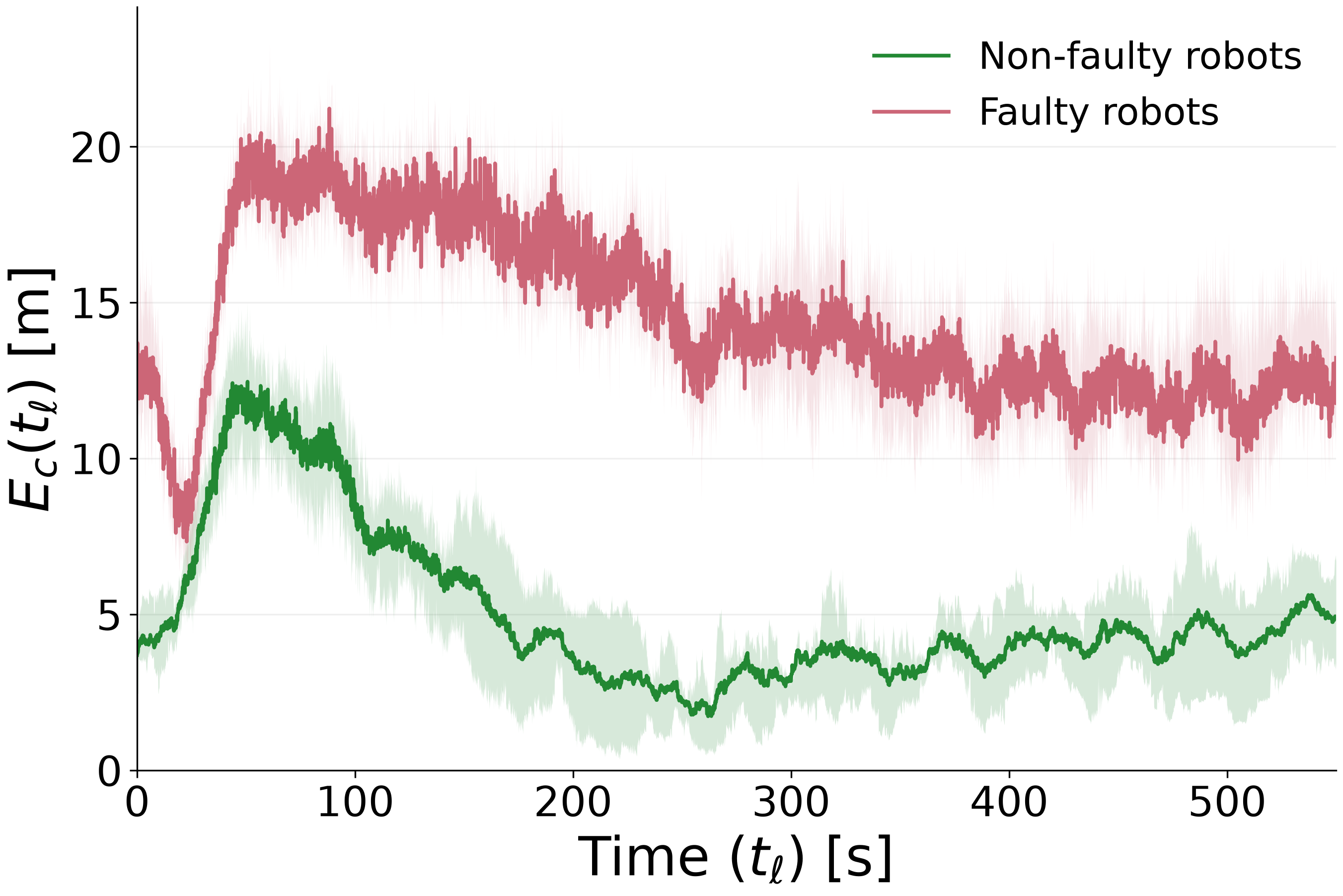}
    }\\
    \subfloat[AOL-C\label{subfig:faulty_time_aolc}]{
    \includegraphics[width=0.75\linewidth]{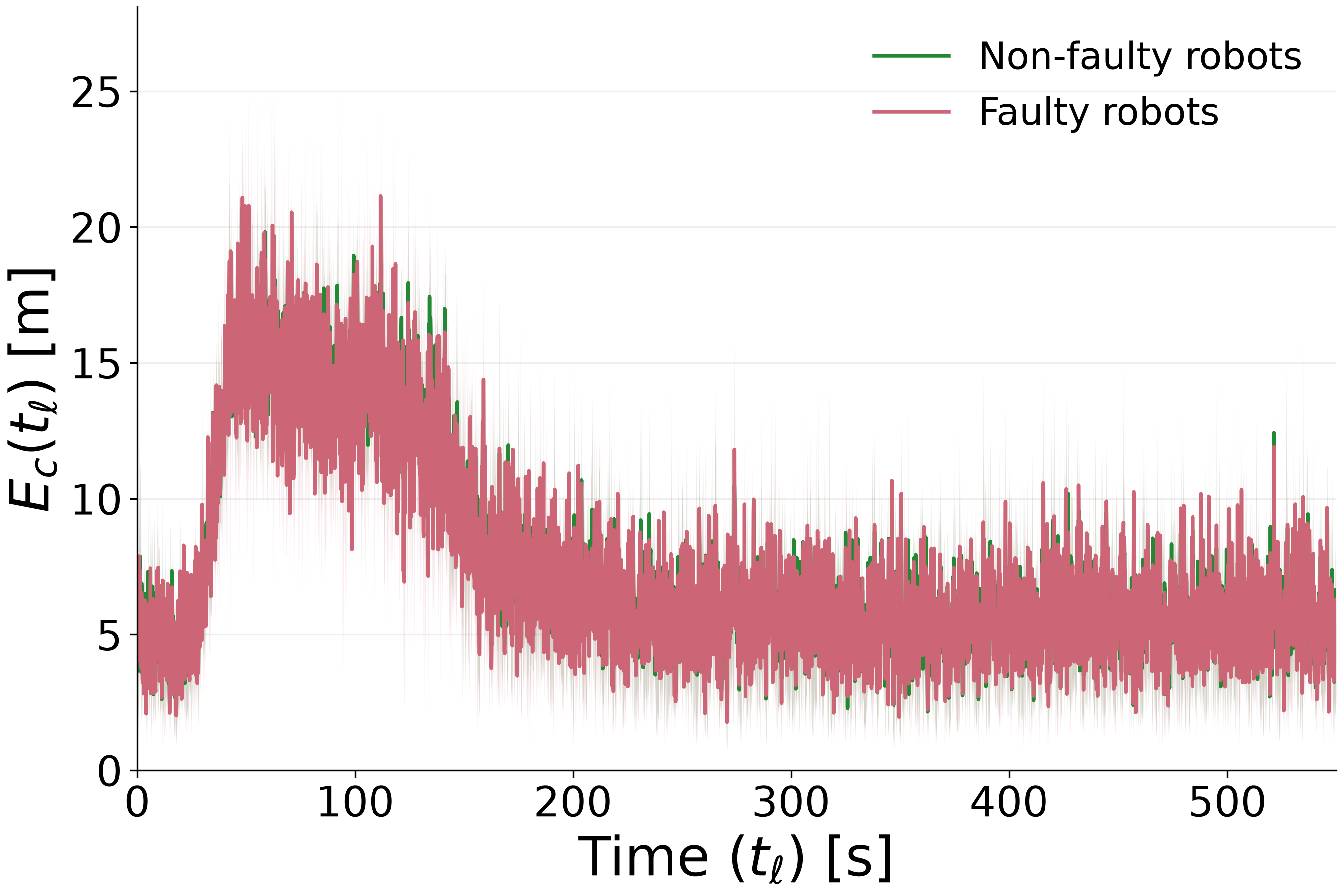}
    }
    \caption{Qualitative comparison of subgroup collective opinion errors (median over evaluation episodes $\ell \in \mathcal{L}$) over time for (a)~DGF-Asym and (b)~AOL-C with \(N=10\), \(\rho=0.4\), and \(\sigma_s^2=0.05\). Errors are evaluated separately for two groups: robots with functioning sensors that can observe the goal (non-faulty, green), and robots with defective sensors or without a goal observation (faulty, red). \textit{Note: The limits of the y-axis vary between plots; this is done for ease of visualization.}
    }
    \label{fig:faulty_time_plot}
\end{figure}

We next compare success rates across a wider range of in-threshold durations~\(\tau\) (Fig.~\ref{fig:persistance_SR_COE}). Both AOL methods outperform DGF-Asym and DGF for short durations, achieving nearly \(100\,\%\) success for \(\tau\leq2\,s\). Their success rates then decline sharply between \(2\) and \(4\,\text{s}\) and approach zero thereafter, indicating that crossings below their respective nominal error thresholds~$E_c^{\mathrm{nom}}$ are predominantly transient. The DGF-based methods show an initial decline in \(SR_{\tau}\), followed by a comparatively stable plateau. This suggests that episodes satisfying the criterion for shorter durations often sustain it over longer intervals. DGF-Asym consistently outperforms DGF, supporting the benefit of reducing the influence of unreliable peers.

For short \(\tau\), the success rates of AOL-C and AOL-1P remain comparatively insensitive to increasing fault ratios~\(\rho\), whereas those of DGF and DGF-Asym decrease progressively (Fig.~\ref{subfig:persistance_SR_COE:a}--\ref{subfig:persistance_SR_COE:c} and \ref{subfig:persistance_SR_COE:d}--\ref{subfig:persistance_SR_COE:f}). This difference may reflect how the methods incorporate unreliable information. AOL-C and AOL-1P synchronously fuse estimates, adapting social influence primarily through distance-based target-detection confidence. Alternating corrections from non-faulty observations and perturbations from faulty ones could explain their high success rates for short \(\tau\) and sharp decline for longer durations. The DGF-based methods instead adapt self-confidence through perceived social influence, reducing reliance on noisy local observations when social information is influential. DGF-Asym additionally weighs peers based on the perceived usefulness of their past estimates. Increasing~\(\rho\), however, reduces the availability of reliable information, potentially limiting error correction and lowering~\(SR_\tau\). Performance deteriorates further with increasing group size at fixed~\(\rho\): success rates decrease (Fig.~\ref{subfig:persistance_SR_COE:d}--\ref{subfig:persistance_SR_COE:f}) and collective opinion error $E_c^\mathrm{con}$ increases (Fig~\ref{subfig:coe_coeJ_N_10},\ref{subfig:coe_noise_N_10}). These results reveal an accuracy-stability trade-off as the fault ratio $\rho$ increases (Figs.~\ref{subfig:coe_coeJ_N_5} and~\ref{subfig:coe_coeJ_N_10}). The DGF-based methods maintain lower collective opinion error jitter than the AOL methods; however, their collective opinion error increases with \(\rho\). DGF-Asym achieves the lowest overall collective opinion error at \(\rho=0.2\), whereas the AOL methods are more accurate at higher fault ratios, particularly for \(N=10\). 

Preliminary experiments show faulty and non-faulty robots forming separate subgroups for DGF-Asym: non-faulty robots converge near the goal and maintain lower opinion errors, while faulty robots remain farther away (Fig.~\ref{subfig:faulty_time_dgfasym}). Qualitative observations suggest that this error gap widens with increasing group size: non-faulty robots remain relatively accurate, whereas faulty robots deteriorate more substantially, contributing to the higher overall $E_c^\mathrm{con}$. Asymmetric peer weighting can limit the influence of faulty estimates, supporting stable, accurate consensus within the non-faulty subgroup without achieving consensus across the entire collective. Under AOL-C, the opinion errors of faulty and non-faulty robots largely overlap and fluctuate substantially (Fig.~\ref{subfig:faulty_time_aolc}). These observations suggest that DGF-Asym can preserve accurate estimates among non-faulty robots without achieving consensus across the entire collective. DGF-Asym's robustness may therefore lie partly in preserving accurate estimates among non-faulty robots, a benefit that collective error and success rate alone may not fully capture.

Across the tested sensor-noise levels, DGF-Asym maintains lower collective opinion error than the AOL baselines (Figs.~\ref{subfig:coe_noise_N_5} and~\ref{subfig:coe_noise_N_10}). Its advantage becomes particularly pronounced at \(\sigma_s^2=25\), where the AOL errors increase substantially. Although DGF-Asym also deteriorates and exhibits greater variability across evaluation periods, its median error remains considerably lower. These results suggest greater robustness to increased sensor noise, despite sensitivity to the proportion of defective robots.

Finally,  the influence threshold heuristic~\(I_{\mathrm{thresh}}\) affects accuracy (Figs.~\ref{subfig:coe_ithresh_N_5} and~\ref{subfig:coe_ithresh_N_10}). 
Among the tested values, \(I_{\mathrm{thresh}}=1\) yields the lowest median error for DGF-Asym with \(N=5\), while \(I_{\mathrm{thresh}}=2\) performs best with \(N=10\). Threshold selection may therefore mitigate the loss of accuracy in larger groups, but no single tested value is optimal across configurations. Since influence accumulates over peer updates, its magnitude depends on the number of communicating peers, their broadcast frequency, and the update interval duration \(T_{\mathrm{up}}\). Developing a normalized or adaptive threshold that accounts for these factors is a direction for future work.

Overall, DGF-Asym produces more stable collective estimates and more reliably satisfies decision criteria requiring sustained accuracy. 
Its accuracy advantage over AOL is strongest under increased sensor noise.
However, it depends on the fault ratio \(\rho\), group size \(N\), and influence threshold~\(I_{\mathrm{thresh}}\).

\section{CONCLUSIONS}

We presented DGF-Asym, a decentralized method for collective estimation under noisy and faulty sensing. 
The method balances personal and social information by using influence-based self-confidence and weights peers using a retrospective measure of predictive performance. 
It supports asynchronous communication without requiring pre-characterized sensor-noise statistics or explicit degradation models. 
Simulation results demonstrate improved stability relative to the baselines across the tested conditions, while revealing sensitivity to fault ratio, group size, and influence threshold.

Retrospective peer weighting assumes that evolving collective opinions generally approach the ground truth, allowing later local estimates to serve as references for evaluating earlier peer information. The influence threshold thus requires configuration-dependent tuning to the expected communication activity within an update interval.
Future work will investigate online threshold adaptation to reduce manual tuning across configurations.

\section*{ACKNOWLEDGMENT}

The authors acknowledge the use of Codex for assistance with the implementation, debugging, and data-analysis scripts.
All AI-assisted code was reviewed, tested, and verified by the authors.


\end{document}